# Air-Releasable Soft Robots for Explosive Ordnance Disposal

Tyler C. Looney,† Nathan M. Savard,† Gus T. Teran,† Archie G. Milligan,† Ryley I. Wheelock,† Michael Scalise,† Daniel P. Perno,†Gregory C. Lewin, Carlo Pinciroli, Cagdas D. Onal, and Markus P. Nemitz*

*Abstract*—The demining of landmines using drones is challenging; air-releasable payloads are typically non-intelligent (e.g., water balloons or explosives) and deploying them at even low altitudes (~6 meter) is inherently inaccurate due to complex deployment trajectories and constrained visual awareness by the drone pilot. Soft robotics offers a unique approach for aerial demining, namely due to the robust, low-cost, and lightweight designs of soft robots. Instead of non-intelligent payloads, here, we propose the use of air-releasable soft robots for demining. We developed a full system consisting of an unmanned aerial vehicle retrofitted to a soft robot carrier including a custom-made deployment mechanism, and an air-releasable, lightweight (296 g), untethered soft hybrid robot with integrated electronics that incorporates a new type of a vacuum-based flasher roller actuator system. We demonstrate a deployment cycle in which the drone drops the soft robotic hybrid from an altitude of 4.5 m meters and after which the robot approaches a dummy landmine. By deploying soft robots at points of interest, we can transition soft robotic technologies from the laboratory to real-world environments.

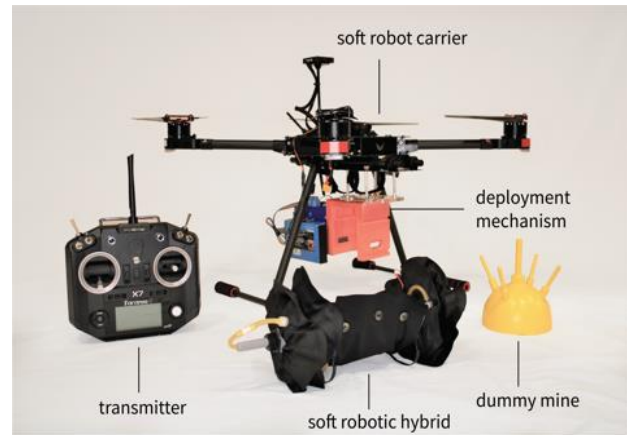

**Figure 1. System overview.** We retrofitted a Tarot 650 drone into a soft robot carrier. A custom-made deployment mechanism can be remote controlled by flicking a toggle switch. The air-releasable soft robotic hybrid employs four vacuum-driven wheels, which dampen the impact from the free fall. We deployed the soft robots successfully from altitudes of 1.5 m, 2.5 m, 3 m, 4.5 m, and 14 m.

## I. INTRODUCTION

There are an estimated 100 million active landmines across the world, impacting the lives of almost a billion people [1]– [3]. Every year, there are approximately 7,000 casualties caused by abandoned, unexploded landmines [4], [5]. Landmines have the potential to seriously injure or kill a person, the severity of the detonation however varies between the types of landmines, with the primary distinction being anti-personnel (AP) and anti-tank (AT) landmines.

AP landmines were originally used to protect AT landmines from being destroyed; however, they have been weaponized to injure victims, rather than killing them, which was developed as a tactic since armies would spend more resources on the injured than on the dead [6]. Improvised explosive devices (IEDs) are also classified as AP landmines but are not industrially manufactured (they typically consist of less or no metal parts), making them often more challenging to detect [7]. AT landmines are larger and require more pressure to detonate than AP landmines because they are designed to damage or destroy tanks or vehicles and those within them [3].

In addition to the 100 million active landmines across the world, 54 countries currently stockpile 180 million AP landmines, and since their planting is easier than their disposal, the development of effective demining strategies remains an important challenge to tackle in the 21st century. It is estimated that a landmine costs between $5 and $30 and the cost of demining a mine is between $300 and $1000.

Attempts to address unexploded landmines have sparked the creation of several methods of Explosive Ordnance Disposal (EOD). Methods of EOD include manual, animal, and machine demining [1].

Manual demining is a process in which human deminers scan the ground with metal detectors and carefully uncover, defuse or destroy mines upon detection [2]. Areas that are being demined typically have metal shrapnel, cans, or other metal objects that can be falsely detected as landmines, or contain plastic landmines that cannot be detected with a metal detector, making manual demining a slow and dangerous procedure [2], [4].

Animal demining utilizes the enhanced sense of smell of animals, such as dogs, rats, or even bees, to locate landmines [2], [7], [9]. The advantage of using animals is that they are not distracted by miscellaneous metal objects; they also have the ability to detect landmines that are not metallic since they are searching for landmines based on odor. Animals are approximately five times faster than humans in finding land mines [1], [2], [7].

Machine demining includes both ground and aerial demining. Ground demining uses heavily armored vehicles or unmanned ground vehicles to detect and detonate landmines. Vehicles typically use one of three demining tools: flails, tillers, or excavators [1], [2]. Flails and tillers utilize counter- directional or co-directional digging to destroy landmines creating pathways in the ground; while excavators dig individual landmines out. Although ground demining may provide more safety for human operators, it constrains demining operations to mostly open terrains.

Aerial demining uses unmanned aerial vehicles (UAVs) to detect and dispose of landmines. UAVs have employed combinations of magnetometers [10], ground penetrating radar [11], or chemical analysis [11], among others; to determine safe pathways through a terrain [12] or to monitor progress in demining operations. UAVs have also been used to detonate mines by dropping heavy payloads or placing explosive payloads that can be remotely detonated [4], [11], [13]. Using UAVs for demining is advantageous because improvements in drone technology allow for rapid scanning speeds of suspicious areas and an increase in safety due to remotely operated systems avoiding contact with soil [11]. Similar to the drawbacks of using drones for commercial delivery, drones for demining suffer from short flight ranges due to limited battery life, can typically carry only small payloads (0.5-3 kg depending on the drone) [10], and drone pilots have limited situational awareness [15]. Payloads also change the center of gravity of the drone which can create bias forces that decrease the stability of the vehicle [16], [17]. This is particularly dangerous when drones carry multiple explosive payloads for demining since a crash could endanger the surroundings.

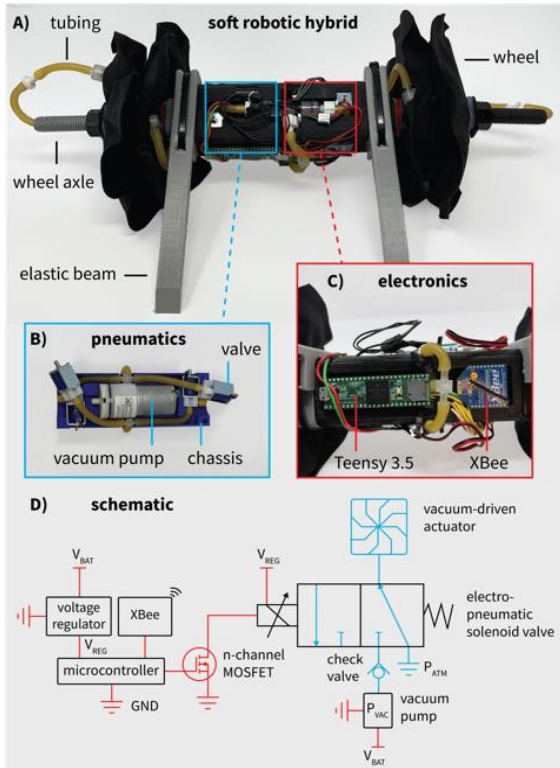

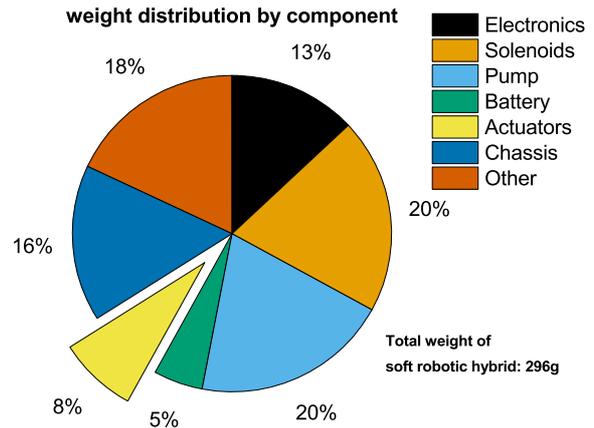

**Figure 2. Untethered soft robotic hybrid.** A) Our soft hybrid consists of electronics, electro-pneumatic components, and four vacuum-driven wheels. We enclosed all components inside a 3D printed chassis. B) The electro-pneumatic components are a vacuum pump, four solenoid valves, and tubing. C) The electronics include a Teensy 3.5 microcontroller, and a XBee communication module to allow for remote controlled motion. D) Schematic of electronics and electro-pneumatics of a single valve and pump.

**Figure 3. Weight distribution of soft robotic hybrid.** Deployable robots need to be carefully optimized for their weight. The heaviest components in our design are the robot chassis, solenoid valves, and a pump, accounting for 56% of the total weight of the robot. The four vacuum-powered actuators only contribute to 8% of the total weight.

The alternative – the detonation of landmines via non-explosive payloads – however, requires highly accurate deployments. In contrast to explosive payloads that have a wide impact area, non-explosive payloads only create forces at their points of impact. In a preliminary study, out of sixteen water balloons deployed from a drone and at an altitude of 6 meters, in ideal weather conditions (low wind speeds), nine water balloons landed within 10 cm of the target [4]. Although the distance between drone and mine can be further reduced to increase deployment accuracy, the likelihood of dispersed projectiles hitting the drone from a detonating landmine increases.

With advances in the relatively new field of soft robotics, there is an emerging alternative to the deployment of non-explosive payloads. Soft materials are known for safety and compatibility with humans and animals [19], [20], low-cost [19], modularity [21], [22], conformability to surroundings [23], high cycle-lifetimes [24]–[26], and damage resistance [27], and which have been often used for interacting with fragile objects such as fruit (i.e., in agriculture), or humans (e.g., in healthcare [17], [18]). They can be purposed for developing soft systems that survive a free fall from a UAV, are able to maneuver post-deployment, and trigger and survive explosions [29]. Instead of deploying non-intelligent water balloons, for the same weight and volume, an air-releasable soft robot can be deployed, which can likely be more effective.

In this work, we take the first step towards a soft robotic demining system and introduce a soft robot carrier including a deployment mechanism, and an air releasable soft robotic hybrid (**Figure 1**). Soft robotic hybrids are robotic systems that employ both, soft and rigid/hard robotic features. The main contributions of this work include:

1. Development of a new actuator system based on pneumatic, artificial muscles; four vacuum-driven wheels are sequentially actuated resulting in net movement.

2. Development of a new type of untethered soft robotic hybrid incorporating our new actuator system.

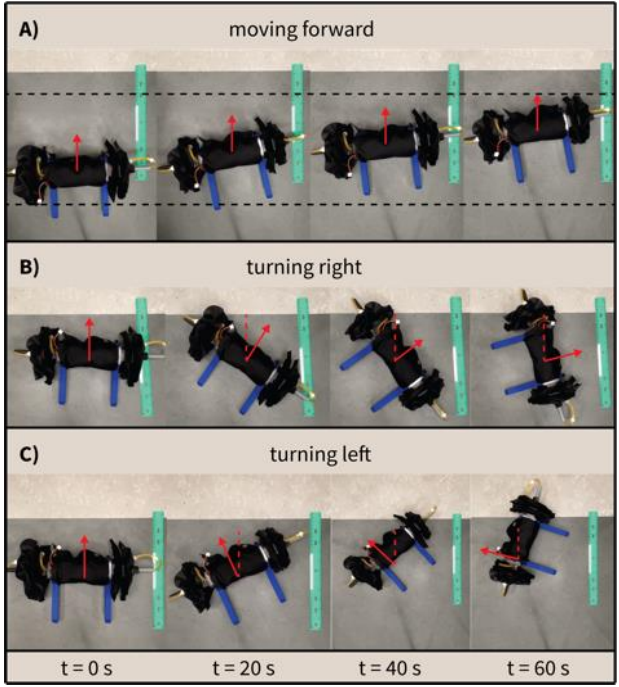

**Figure 4. Movement patterns using vacuum-driven wheels.** The four wheels can be actuated in three different patterns. A) The inner wheels are actuated first, followed by the outer wheels, allowing for forward locomotion. Then the inner wheels are exposed to atmospheric pressure first, followed by the outer wheels, allowing for forward locomotion again and returning to the original configuration. B) A combination of inner and outer wheels are actuated first, followed by the other pair, allow for turning right. C) Similar to B but reversed and leading to a left instead of a right turning locomotion.

3. Development of an aerial soft robot carrier including a remote-controlled deployment mechanism.
4. Demonstration of the soft robot carrier deploying our soft robotic hybrid in proximity to a dummy mine, and the soft robotic hybrid being remote controlled towards the mine.

## II. DESIGN

### A. Kirigami-based, vacuum-driven flasher roller pattern

We developed an actuator system that is comprised of four vacuum-driven wheels and based on previous work on fluid-driven, origami-inspired artificial muscles (**Figure 2**) [30]. We chose fluid-driven actuators over electric motors and conventional wheels to reduce the risk of the robot breaking. Our vacuum-driven wheels are expanded in the unactuated state and dampen the impact of a free-fall (**Movie S1**). Each wheel consists of a polyethylene sheet cut into a kirigami pattern with an outer shape of a square and sealed in a thermoplastic coated textile. When a vacuum of 55 kPa is applied to the wheel, it contracts (reducing in size from approximately 166 cm$^2$ to 110 cm$^2$, or 34% contraction) and twists by 45 degrees. We created two types of wheels, the CW wheel twists c̲lockw̲ise when a vacuum is applied, and the CCW wheel twists counter-c̲lockw̲ise when a vacuum is applied. Our actuator system consists of two CW and two CCW wheels attached to a robot axle. This configuration allows us to create asymmetric motion that results in net movement. We developed three actuation patterns that allow the robot to move straight and take left and right turns (**Figure 4**).

The motion of our robot is defined by the action of the largest wheel, i.e., either the last wheel to contract due to an applied vacuum or the first wheel to expand due to releasing the vacuum. Contracting and expanding wheels in certain orders creates three different types of motion: moving forward, turning left, and turning right. We denote the wheels (1) to (4) from the left to the right (**Figure 2**). Hence, (1) and (4) are the outer wheels (CW), and (2) and (3) the inner wheels (CCW). Each wheel is either applied to vacuum ($P_{1-4}$= -55 kPa) or atmospheric pressure ($P_{1-4}$= $P_{atm}$), causing contraction, or expansion to its original shape.

Moving forward: In the initial state, all wheels are exposed to atmospheric pressure ($P_1$=$P_2$=$P_3$=$P_4$= $P_{atm}$). The inner wheels are applied to vacuum ($P_2$=$P_3$= -55 kPa), and once they are contracted, the outer wheels are applied to vacuum ($P_1$=$P_4$= -55 kPa). With wheels (1) and (4) contracted last, the robot will generate an overall forward motion. Instead of resetting the wheels to atmospheric pressure and repeat the actuation pattern, we expose the inner wheels to atmospheric pressure first, generating a forward movement due to their expansion ($P_2$=$P_3$= $P_{atm}$). We then expose the outer wheels to atmospheric pressure returning to the original configuration ($P_1$=$P_2$=$P_3$=$P_4$= $P_{atm}$). With this method the robot achieved maximum velocities of 12 cm/min.

Turning left: The wheels start in the idle state ($P_1$=$P_2$=$P_3$=$P_4$= $P_{atm}$). Vacuum is applied to the outer wheel and inner wheel ($P_1$=$P_3$= -55 kPa) and once they are contracted, vacuum is applied to the other two wheels ($P_2$=$P_4$= -55 kPa). Because (2) is a CCW and (4) a CW wheel, the robot generates a turn motion to the left. We achieved maximum rotational velocities of 70 deg./min. Similar to our forward motion,

Turning right: The wheels start again in idle state ($P_1$=$P_2$=$P_3$=$P_4$= $P_{atm}$), vacuum is applied to an outer and an inner wheel ($P_2$=$P_4$= -55 kPa), and finally applied to vacuum ($P_1$=$P_3$= -55 kPa). Because (1) is a CW and (3) a CCW wheel, the robot generates a turn motion to the right.

All locomotion requires an expenditure of energy to overcome the resistance to motion, namely, friction. In our case, our robot is lightweight (296g) and depending on the contact surface between robot and environment, friction is minimal, and the wheels can slip. To prevent the robot from slipping, we added flexible beams to the robot that make firm contact with the ground (**Figure 2**), acting as an asymmetric friction connection, allowing forward, but preventing backward movements.

The CW and CCW wheel patterns were cut from a polyethylene sheet with a thickness of 0.15 mm, into a total of four squares with a side length of 129 mm, and one sheet for each wheel, using a laser cutter (Universal Laser Systems PLS6.150D). After the squares were cut, they were folded via mountain and valley folds to create the 3D contractable and rotatable pattern shapes. The outer body was created from rectangular thermoplastic polyurethane (TPU) coated nylon textile patches (Seattle Fabrics Inc.). We placed the TPU coated textile patch on the bottom side of the heat press, a parchment paper was laid on top of the patch. The textile patch was then folded over the parchment paper, allowing the parchment to extend from one side of the heat press to form a pocket. A heat press (Fancierstudio) was then used at 220 °C for 35 seconds to bond the TPU coated sides of the textile on three sides. The polyethylene pattern was placed inside the textile pocket along with a pneumatic port (Luer-Lock connector) and a 3D printed spacer before being sealed with the heat press. We supply the design files on our webpage ([www.roboticmaterialsgroup.com/tools](www.roboticmaterialsgroup.com/tools)).

### B. Untethered, soft robotic hybrid

We incorporated the actuation system in an untethered soft robotic hybrid that also holds the electronics and a Lithium Polymer battery (**Figure 2**).

The electronics consists of a microcontroller (Teensy 3.5), a buck converter (Drok DC 4.5V~24V to 5V 3A), five MOSFETs (Nxperia PSMN017-30PL), an XBee communication module, four solenoid valves (Adafruit 4663 6V Air Valve), a micropump (Adafruit 4699 4.5V), and a Lithium Polymer battery (2S 7.4V 200mAh). Our microcontroller controls the MOSFETs; the MOSFETs switch on and off the solenoid valves and the micropump. The microcontroller receives commands (move straight, turn left, or turn right) via ZigBee from a laptop. The total power consumption of the robot is 8.8 W when moving, and 0.35 W when in an idle state (microcontroller and ZigBee communication operating), allowing for ~10 minutes of motion and ~4 hours of idle state (**Figure 2**).

We designed a rectangular prism body to hold the components (**Figure 2**) and which can be securely loaded into a deployment mechanism. This body was 3D printed using PLA at 10-15% gyroid infill pattern. We designed two endcaps with hollowed threaded axles to allow a silicone tube with an inner diameter of 1.6 mm and an outer diameter of 4.7 mm to be integrated. These threaded axles were also printed using PLA with a 15% gyroid infill and allow the robot wheels to be secured in place with the use of 3D printed spacers and washers, and nylon nuts. We tightened the endcaps onto the central body via four screws for each endcap.

We installed check valves between solenoid valves and micropump, to prevent airflow from passing between wheels connected to atmospheric pressure to wheels that are under vacuum (**Figure 2**)

The microcontroller runs a simple control loop that awaits a command from the XBee module using an UART Serial connection. The XBee module receives the command as a character from a paired XBee module that is connected to a computer. Once the robot receives the character command, it executes the corresponding motion pattern which is stored in a look-up table. Our robot completes motion patterns by actuating the vacuum-driven wheels using solenoids. The XBee has a hardware queue that allows the computer to send a queue of motion commands. The actuation cycles will be completed before awaiting another command.

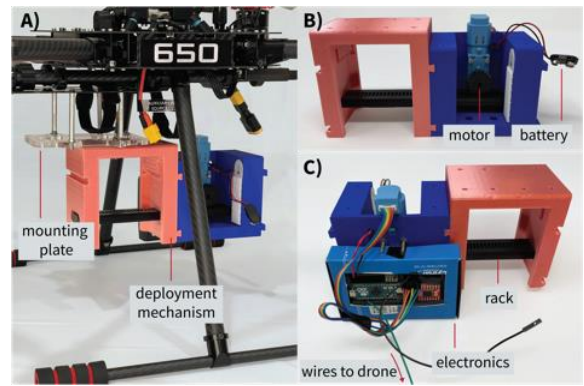

**Figure 5. Custom 3D printed deployment mechanism.** A) We attached our deployment mechanism via a mounting plate to the drone. B) The deployment mechanism uses an electric motor with integrated encoder and is powered by a 9V battery. C) A microcontroller is connected to the flight controller of the drone and controls the motor using a motor control board.

### C) Soft robotic carrier and deployment mechanism

Our soft robot carrier is a Tarot 650 drone with a custom-made deployment mechanism. It is powered by an 8000 mAh battery, allowing a total flight time of 14 minutes with a payload of 1.5 kg, and 23 minutes without. It is equipped with a mRo-X2.1 flight controller and is controlled with a 2.4 GHz transmitter (Taranis QX7 ACCST). We operate ArduPilot on the flight controller of our drone including a flight stabilizer, allowing the drone to hover automatically at constant altitude. We used Mission Planner (software) to configure the drone, monitor the status of the drone, and view flight logs. Mission Planner can also be used to let the drone autonomously fly to the location of a mine, deploy the soft robot, and return home; a feature that we plan on using in future work. Several authors are certified drone pilots complying with Part 107 of the FAA regulations and followed WPI protocols when flying on university premises.

The deployment system is based on a rack and pinion mechanism (**Figure 5**) and was 3D printed from PLA using a Prusa MK3S**.** A microcontroller (Arduino MICRO) controls an electric motor with integrated encoder (DG01D-E) via a motor driver (TB6612FNG). The motor rotates a circular gear that engages the rack (linear gear), and slides it to the side, releasing the soft robotic hybrid. We interfaced our microcontroller with the flight controller of the drone by configuring a GPIO pin of the flight controller as a relay pin. The GPIO pin of the flight controller outputs a low voltage to the microcontroller when a toggle switch of the drone transmitter is flipped. This setup allows the drone pilot to release a soft robot midair at the flick of a switch.

The deployment mechanism weighs 320 g. We supply the design files of the deployment mechanism on our webpage.

### III. DEMONSTRATION

We demonstrated our entire system on WPI's Alumni Field (**Figure 6**). The drone carried the soft robotic hybrid and approached our dummy landmine (**Figure 6A**). Once it was in close proximity to the landmine (**Figure 6B**), we engaged the deployment mechanism and dropped the soft robotic hybrid from an altitude of 3.5 meters. After deployment, we safely landed the soft robot carrier (**Figure 6C**). We used then remote control to move the soft robotic hybrid the remaining meter to the landmine (**Figure 6D**). We performed a total of 8 tests, deploying the robot from a range of altitudes, from 1 meter to 14 meters without breaking it. The vacuum-driven wheels were expanded (connected to atmospheric pressure) during deployment, acting as cushions and dampening the impact from the free fall.

### IV. DISCUSSION

In this paper, we introduce a soft robot carrier, a deployment mechanism for soft robots, and an air-releasable soft robotic hybrid. The soft robotic hybrid employs a new type of actuation mechanism, a vacuum-based flasher roller.

#### A. Importance of teams of conventional and soft robots

Soft robots are often perceived as laboratory-bound curiosities with a vast potential for real-world applications. Compared to wheeled mobile robots, soft robots lack fast actuation speeds, making them non-optimal for many applications, including search and rescue or demining. Our soft robotic hybrid, although untethered, is also slow (12 cm/min and 70 deg/min.).

However, there is a vast potential for robot teams consisting of conventional and soft robots. While conventional robots such as UAVs, possess fast actuation speeds, they are fragile to interactions with the environment. The qualities and disadvantages of soft robots are completely opposite to conventional robots. Soft robots are slow, but robust and dynamic in their interaction with the environment. By combining conventional and soft robotic technologies, we can mitigate the drawbacks of each technology and take advantage of their combined abilities.

#### B. Using soft robots for demining

Soft robots are uniquely suited for demining purposes. While we introduce a complete framework (i.e., the soft robot carrier) for the development of deployable soft robots, our soft robotic hybrid depicts a proof-of-concept only. Work from Shepherd et al. has shown that soft robots can use combustion for locomotion, and it has also shown that soft robots can be designed to *survive* explosions [29]. Therefore, we anticipate a research direction in which untethered robots move onto mines, initiate an internal explosion that triggers the landmine with the potential of the soft robot surviving the detonation.

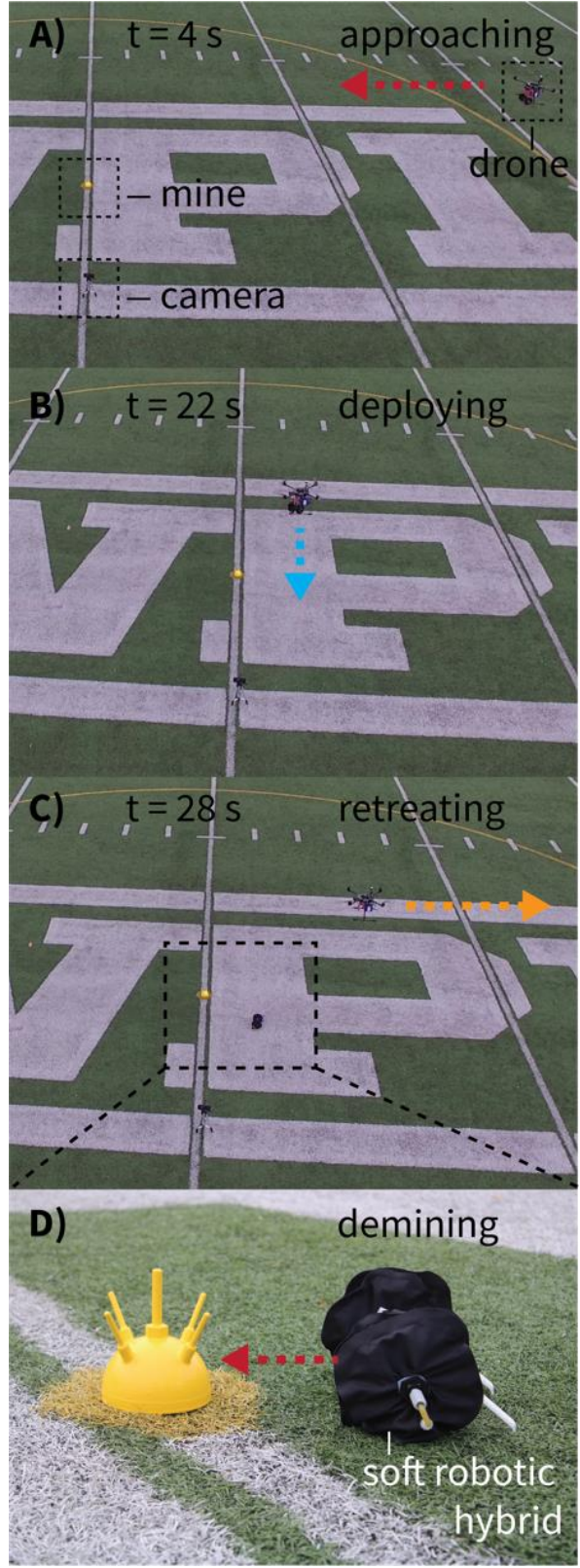

**Figure 6. Demonstration of a deployment cycle.** A) The soft robot carrier departs and heads towards the dummy landmine. B) Once it is in proximity to the mine, it deploys the soft robotic hybrid. C) The soft robot carrier returns, and (D) the soft robotic hybrid approaches the landmine.

## C. Important abilities of air-releasable soft robots

Our demining project has taught us several lessons in the development of air-releasable soft robots. First, soft robots can be designed to be lightweight; our actuators are only 8% of the overall weight of our untethered soft robots (**Figure 3**). Most of the heavy components are electrical-based (valves, pumps). Therefore, we continue research on developing soft analogues for electrical (not electronic) components. Second, the ability of soft robots to fold into small sizes is extremely important when being deployed from a drone. A drone is not only limited in the maximum weight, but also the maximum volume it can carry. Physically large robots increase the surface area of the drone and impact its flight behavior and control.

## V. CONCLUSION

The use of conventional robots for the deployment of soft robots has far reaching implications and pushes soft robotics research into the real world. Drones can be retrofitted to soft robot carriers to deploy soft robots at points of interest. A deployment mechanism can be easily 3D printed and, using simple electronics, can be connected to the flight controller of the drone, to be operated with the drone transmitter. Untethered soft robots are mostly bulky and heavy because of their valves and pump, encouraging further research on soft substitutes for electro-pneumatic components.

## VI. ACKNOWLEDGEMENT

We thank Malek ElShakhs, Alexander Hagedorn, Tory Howlett, Eamon Oldridge, Maggie Raque, and Jeremy Wong for their previous efforts on the Demining Major Qualifying Project (MQP) at WPI; Brian Katz and Katelyn Wheeler for their help creating prototypes of kirigami wheels; Brandon Simpson and Cole Parks for lending us their 3D printers; and Nicole Kuberka for supplying us with electronics equipment and materials. A special thanks goes to Professors Craig Putnam, Nicholas Bertozzi, and Ken Stafford for their support of previous MQP teams.

## REFERENCES

[1] D. Mikulic, "Humanitarian Demining Techniques," in *Design of Demining Machines*, Springer London, 2013, pp. 1–27. doi: 10.1007/978-1-4471-4504-2_1.

[2] M. Habib and M. K. Habib, "Mine Clearance Techniques and Technologies for Effective Humanitarian Demining," Maki, 2002.

[3] J. Trevelyan, *Humanitarian Demining : Research Challenges*.

[4] M. Elshakhs et al., "Demining Autonomous System A Major Qualifying Project," 2021. [Online]. Available: https://www.wpi.edu/academics/undergraduate

[5] M. Armstrong, "Casualties of Landmines [Digital image]," https://www.statista.com/chart/20679/casualties-of-landmines-timeline/, Jan. 31, 2020.

[6] K. Ehlers et al., "Demining Autonomous System A Major Qualifying Project." [Online]. Available: https://www.wpi.edu/academics/undergraduate

[7] P. A. Prada and M. Chávez Rodríguez, "Demining Dogs in Colombia - A Review of Operational Challenges, Chemical Perspectives, and Practical Implications," *Science and Justice*, vol. 56, no. 4. Forensic Science Society, pp. 269–277, Jul. 01, 2016. doi: 10.1016/j.scijus.2016.03.002.

[8] R. J. Swinton and D. M. Bergeron, "Evaluation of a Silent Killer, the PMN Anti-Personnel Blast Mine," 2004.

[9] J. Filipi et al., "Honeybee-based biohybrid system for landmine detection," *Science of the Total Environment*, vol. 803, Jan. 2022, doi: 10.1016/j.scitotenv.2021.150041.

[10] L. S. Yoo, J. H. Lee, S. H. Ko, S. K. Jung, S. H. Lee, and Y. K. Lee, "A Drone Fitted with a Magnetometer Detects Landmines," *IEEE Geoscience and Remote Sensing Letters*, vol. 17, no. 12, pp. 2035–2039, Dec. 2020, doi: 10.1109/LGRS.2019.2962062.

[11] M. G. Fernandez et al., "Synthetic aperture radar imaging system for landmine detection using a ground penetrating radar on board a unmanned aerial vehicle," *IEEE Access*, vol. 6, pp. 45100–45112, Aug. 2018, doi: 10.1109/ACCESS.2018.2863572.

[12] A. W. Dorn, "Eliminating hidden killers: How can technology help humanitarian demining?," *Stability*, vol. 8, no. 1, pp. 1–17, 2019, doi: 10.5334/sta.743.

[13] "Mine Kafon Airborne Demining System."

[14] A. Karak and K. Abdelghany, "The hybrid vehicle-drone routing problem for pick-up and delivery services," *Transportation Research Part C: Emerging Technologies*, vol. 102, pp. 427–449, May 2019, doi: 10.1016/j.trc.2019.03.021.

[15] T. Kille, P. R. Bates, S. Y. Lee, and D. M. Kille, "Situational awareness training for operators of unmanned aerial vehicles," *Proceedings of the 21st International Symposium on Aviation Psychology (ISAP 2021). International Symposium on Aviation Psychology*, 2021.

[16] IEEE Staff, *2017 14th International Conference on Ubiquitous Robots and Ambient Intelligence (URAI)*. IEEE, 2017.

[17] P. E. I. Pounds, D. R. Bersak, and A. M. Dollar, "Stability of small-scale UAV helicopters and quadrotors with added payload mass under PID control," *Autonomous Robots*, vol. 33, no. 1–2, pp. 129–142, Aug. 2012, doi: 10.1007/s10514-012-9280-5.

[18] P. Polygerinos et al., "Soft Robotics: Review of Fluid-Driven Intrinsically Soft Devices; Manufacturing, Sensing, Control, and Applications in Human-Robot Interaction," *Advanced Engineering Materials*, vol. 19, no. 12. Wiley-VCH Verlag, Dec. 01, 2017. doi: 10.1002/adem.201700016.

[19] P. Polygerinos, Z. Wang, K. C. Galloway, R. J. Wood, and C. J. Walsh, "Soft robotic glove for combined assistance and at-home rehabilitation," in *Robotics and Autonomous Systems*, Nov. 2015, vol. 73, pp. 135–143. doi: 10.1016/j.robot.2014.08.014.

[20] R. M. McKenzie, M. E. Sayed, M. P. Nemitz, B. W. Flynn, and A. A. Stokes, "Linbots: Soft Modular Robots Utilizing Voice Coils," *Soft Robotics*, vol. 6, no. 2, pp. 195–205, Apr. 2019, doi: 10.1089/soro.2018.0058.

[21] M. P. Nemitz, P. Mihaylov, T. W. Barraclough, D. Ross, and A. A. Stokes, "Using Voice Coils to Actuate Modular Soft Robots: Wormbot, an Example," *Soft Robotics*, vol. 3, no. 4, pp. 198–204, Dec. 2016, doi: 10.1089/soro.2016.0009.

[22] F. Ilievski, A. D. Mazzeo, R. F. Shepherd, X. Chen, and G. M. Whitesides, "Soft robotics for chemists," *Angewandte Chemie - International Edition*, vol. 50, no. 8, pp. 1890–1895, Feb. 2011, doi: 10.1002/anie.201006464.

[23] P. Rothemund et al., "A soft, bistable valve for autonomous control of soft actuators," 2018. [Online]. Available: https://www.science.org

[24] D. Yang et al., "Buckling Pneumatic Linear Actuators Inspired by Muscle," *Advanced Materials Technologies*, vol. 1, no. 3, Jun. 2016, doi: 10.1002/admt.201600055.

[25] D. J. Preston et al., "A soft ring oscillator," 2019. [Online]. Available: https://www.science.org

[26] M. T. Tolley et al., "A Resilient, Untethered Soft Robot," *Soft Robotics*, vol. 1, no. 3, pp. 213–223, Sep. 2014, doi: 10.1089/soro.2014.0008.

[27] M. Runciman, A. Darzi, and G. P. Mylonas, "Soft Robotics in Minimally Invasive Surgery," *Soft Robotics*, vol. 6, no. 4, pp. 423–443, Aug. 2019, doi: 10.1089/soro.2018.0136.

[28] J. A. Blaya and H. Herr, "Adaptive Control of a Variable-Impedance Ankle-Foot Orthosis to Assist Drop-Foot Gait," *IEEE Transactions on Neural Systems and Rehabilitation Engineering*, vol. 12, no. 1, pp. 24–31, Mar. 2004, doi: 10.1109/TNSRE.2003.823266.

[29] R. F. Shepherd et al., "Using explosions to power a soft robot," *Angewandte Chemie - International Edition*, vol. 52, no. 10, pp. 2892–2896, Mar. 2013, doi: 10.1002/anie.201209540.

[30] S. Li, D. M. Vogt, D. Rus, and R. J. Wood, "Fluid-driven origami-inspired artificial muscles," *Proceedings of the National Academy of Sciences of the United States of America*, vol. 114, no. 50, pp. 13132–13137, Dec. 2017, doi: 10.1073/pnas.1713450114.